\documentclass{svproc}
\usepackage{url}

\usepackage{graphicx}
\usepackage{xcolor}
\usepackage{comment}

\usepackage{subfig}
\usepackage{caption}
\usepackage[export]{adjustbox}
\usepackage{multirow}
\usepackage{multicol} 
\usepackage{footmisc}

\begin{document}
\mainmatter              
\title{Accelerating Road Sign Ground Truth Construction with Knowledge Graph and Machine Learning}
\titlerunning{Road Sign Ground Truth}  
%
%
\authorrunning{Kim et al.} 
%
%
\author{Ji Eun Kim$^*$, Cory Henson$^*$, Kevin Huang$^*$, Tuan A. Tran$^+$, Wan-Yi Lin$^*$}

\institute{$^*$Bosch Corporation Research, $^+$Bosch Chassis Systems Control, \\
\email{\{jieun.kim, cory.henson, kevin.huang,anhtuan.tran2, wan-yi.lin\}@bosch.com}
}


\maketitle              
\begin{abstract}
 Having a comprehensive, high-quality dataset of road sign annotation is critical to the success of AI-based Road Sign Recognition (RSR) systems. 
 In practice, annotators often face difficulties in learning road sign systems of different countries; hence, the tasks are often time-consuming and produce poor results.
 We propose a novel approach using knowledge graphs and a machine learning algorithm - variational prototyping-encoder (VPE) - to assist human annotators in classifying road signs effectively. Annotators can query the Road Sign Knowledge Graph using visual attributes and receive closest matching candidates suggested by the VPE model. The VPE model uses the candidates from the knowledge graph and a real sign image patch as inputs. We show that our knowledge graph approach can reduce sign search space by 98.9\%. Furthermore, with VPE, our system can propose the correct single candidate for 75\% of signs in the tested datasets, eliminating the human search effort entirely in those cases. 
 

\end{abstract} 


\section{Introduction}
Recognizing and understanding road signs are important features of advanced driver-assistance systems (ADAS), which are offered in modern vehicles via technologies such as road-sign recognition (RSR) or intelligent speed adaption (ISA)\footnote{These features are mandatory in all new cars sold within Europe from May 2022~\cite{EU_2019_2144}}. Recent RSR and ISA solutions heavily use Machine Learning methods and require comprehensive, high-quality datasets of road sign annotation as ground truth. To be ready for real-world usage, the ground truth must be built from test drives around the world. The number of road sign images to be annotated can be enormous, up to more than ten millions each year.
Any representative sample of these images that covers enough countries and conditions will be of considerable size. It is therefore crucial to optimize the annotation task and minimize annotator's time in each session.

There are two main challenges in performing a road sign annotation task. First, there are too many road signs to search through to find a matching one (USA alone has more than 800 federally approved road signs, and 10 states in USA have their own state conventions which are different from the federal convention \cite{mutcd2006manual}). This makes manual classification of each sign instance against a full palette of signs infeasible\footnote{Note that we define "sign" as a prototypical sign, and differentiate it from "sign instance" which is an instance of a prototypical sign as seen in drive data. For example, if a set of images contains 4 stop signs and 4 yield signs, then there are 8 sign instances in total, and 2 signs (stop and yield).}. One solution is to have a machine learning system limit the number of candidates for human annotators to search from (e.g., to 5 signs). 

\begin{figure}[hbt!]
  \includegraphics[width=0.95\textwidth]{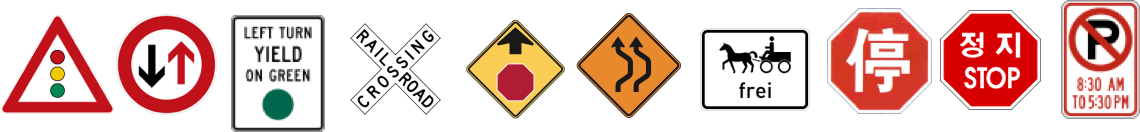}
  \caption{Examples of Various Road Signs Across Countries Representing Different Shapes, Colors, Icons and Languages}
  \label{fig:signs_variation}
\end{figure}

The second challenge lies in the fact that different countries follow different conventions on road signs. For instance, USA follows MUTCD~\cite{mutcd2006manual}, while European countries adopt the Vienna convention~\cite{inland1968convention}. Some countries adopt multiple conventions, and some introduce different variants in features such as colors, fonts, size, etc. This is illustrated in Figure \ref{fig:signs_variation}. No annotator possesses full knowledge of all road sign systems and may choose the wrong ones, especially when the instance is not clear (e.g., gray-scale images, night images, and so forth).

In this work, we address the above issues with a novel solution that combines knowledge graph and machine learning to assist annotators and accelerate the ground truth annotation. The idea is that all road signs have some basic visual features, and we can navigate through the knowledge graph of these visual features (focusing on country-specific sub-graphs by using GPS data associated with the images), to locate the candidate signs, supporting the sense-making of human annotators.
We show that this approach can reduce the search space for signs by 98.9\%, from 845 signs to 8.92 signs. This reduction in search space translates to reduced search effort and time by human annotators for locating the correct sign.
To further reduce the search space, we propose to use a Variational Protoyping-Encoder (VPE~\cite{vpe2019}) that uses one-shot learning to find matching signs, even if unseen in prior training data. To summarize our contribution, in this work we have:
\begin{itemize}
  \item introduced Road Sign Ontology (RSO) to represent salient features of road signs, 
  \item proposed crowd-sourcing techniques to construct the Road Sign Knowledge Graph at scale across countries and states
  \item used knowledge graph to reduce the sign search space by 98.9\%
  \item built a VPE model that is combined with the knowledge graph to further reduce the search space. This approach can propose the correct single candidate for 75\% of signs in the tested datasets even for unseen signs during the training phase, eliminating the human search effort altogether in those cases
\end{itemize}

In the following sections, we quickly summarize related work, explain the construction of the ontology and knowledge graph, and detail our knowledge graph and machine learning-assisted road sign annotation pipeline. Finally, we present our experimental results on internal and public datasets.
\section{Related Work}
\paragraph{Road Sign Detection and Classification}
Due to the ubiquitous presence of road signs, detecting and classifying them are important tasks in automated driving research~\cite{zhu2016traffic,gtsrb2011}. A typical framework involves the detecting of bounding boxes containing road signs from video or image data~\cite{lisa2012}, then mapping them to canonical signs, often represented by an image prototype. While much of existing work focused on the visual properties of the road signs such as color~\cite{lopez2007color}, shape, and templates~\cite{loy2004fast,malik2007road}, the varying of such properties as regulated in different conventions is not well addressed~\cite{lisa2012}. 
Although there are several open datasets for road sign recognition and classification~\cite{gtsrb2011,timofte2014multi,larsson2011using,belaroussi2010road}, they cover only a very small portion of the signs and face several challenges when adapted to the international setting. 

\paragraph{Using Knowledge Graph to Improve Evaluation}
Our work is characterised by the usage of an ontology and knowledge graphs to assist human annotators in complex tasks. In this perspective, it has been studied in crowd-sourcing research communities, by introducing domain knowledge to support human sense-making and learning during the microtask.
Early work in this direction can be found in database or text mining applications~\cite{chu2015katara,dojchinovski2016crowdsourced}. For visual tasks, some open ontologies are built to help humans annotate the biomedical concepts~\cite{bukhari2015bim}. To the best of our knowledge, the use of a large-scale knowledge graph, consisting not only of taxonomy but also of Resource Description Framework (RDF) facts curated from several sources for evaluating visual tasks in autonomous driving area is completely unexplored. 

Our construction of the Road Sign Knowledge Graph is also inspired by related work in evaluating quality of large-scale linked data using crowd-sourcing~\cite{jou2015visual,simperl2013cldqa}. For images, previous work suggests that careful design of microtask should take into account visual effects on
human across countries~\cite{jou2015visual}.

\paragraph{Meta-Learning for Road Signs}
A recent approach for dealing with the incompleteness of road sign classes is to use meta-learning, including few-shot and one-shot learning~\cite{zhou2019few,vpe2019}. The idea is to represent each unseen road sign class with an image prototype, which is jointly encoded with existing classes, either driven by a distance metric or a meta-model, to learn to translate each real image depicting a road sign. Some work combines knowledge graph into meta-learning to improve the performance via concept propagation~\cite{kampffmeyer2019rethinking}. In autonomous driving, to the best of our knowledge, our work is the first to propose the combination of knowledge graph and one-shot learning to construct ground truth efficiently.

\section{Road Sign Ontology} 
\label{sec-rso}
\begin{figure}[ht!]
    \centering
    \subfloat[Class Hierarchy]{{\includegraphics[width=4.2cm]{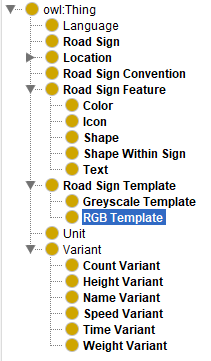} }}%
    \qquad
    \subfloat[Object Properties]{{\includegraphics[width=5.2cm]{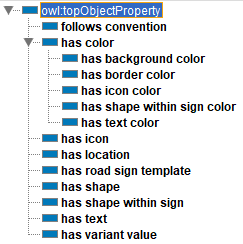} }}%
    \caption{Visualization of the Road Sign Ontology with Protege \cite{protege}}%
    \label{fig:rs-ontology}%
\end{figure}

This section describes an ontology for representing and reasoning about road signs; namely, the Road Sign Ontology (RSO). This ontology, and its conformant knowledge graph (see Section 4), is used to assist in the data annotation process and the training of machine learning models for road sign classification. RSO seeks to represent the salient visual features of a road sign that are discernible through sight or imaging, and is modeled using the Web Ontology Language (OWL2) \cite{owl2}. See Figure \ref{fig:rs-ontology} for a visualization of the primary ontology concepts.

While designing RSO, two primary requirements were considered. First, the ontology should represent the features of road signs that are beneficial to the performance of machine learning algorithms.
Second, the ontology should represent concepts at an appropriate level of granularity that enables human annotators to effectively identify road signs and their visual features when looking at an image.  

\subsection{Road Sign Features}
The primary features of a road sign to be represented include its shape, color, text, and printed icons. 

\paragraph{Shape} RSO distinguishes between two types of shapes associated with a road sign. The most obvious is the shape of the physical plate. For example, in the United States, stop signs have an octagon shape, yield signs have a downward triangle shape, and speed limit signs have a rectangular shape. There are 11 different shapes that the physical plate of a road sign could take. The second type of shape includes geometric shapes that are printed on the physical plate. Common printed geometric shapes include arrows, circles, and diagonal lines. RSO represents 9 different printed shapes.

\paragraph{Color} Similar to shape, RSO distinguishes between multiple different types of color associated with a road sign. Specifically, a road sign can have a foreground color, a background color, and a border color. 11 common colors are enumerated within the ontology.
    
\paragraph{Icon} Icons are a special type of shape printed on a road sign that depict various objects. The types of objects often depicted include vehicles, people, animals, and assorted traffic infrastructure (e.g., a traffic light). Given the large number of possible distinct icons, RSO only defines a few common categories, including: animal, infrastructure, nature, person, vehicle, and other.     
    
\paragraph{Text} Many road signs include printed text. Stop signs print the word \textit{STOP}, yield signs print the word \textit{YIELD}, and speed limit signs include both the words \textit{Speed Limit} and a number. Rather than enumerating all possible text that may be printed on a sign, RSO allows the text of a specific sign to be annotated using an OWL Datatype Property. While RSO does not define enumerations for all possible text on a road sign, it does enable the categorization of text into various types, based on the intended meaning or use. The categories of text include: speed, height, weight, time, name, and number. As an example, the text of a speed-limit sign is identified with the \textit{speed} category, while the text of a sign announcing entrance to a town is identified with the \textit{name} category.


\subsection{Road Sign Conventions and Prototypes}
In practice, road signs must adhere to convention. It is this adherence that allows and empowers a person to detect and identify the meaning of a sign with only a brief glance. Such conventions define rules and constraints on how road signs of various types should be printed and displayed.
There are three primary road sign conventions in use today: the Vienna Convention \cite{united2007convention}, the MUTCD Convention \cite{mutcd2006manual}, and the SADC Convention \cite{sa_url}. The Vienna Convention is used primarily in Europe and China; the MUTCD Convention (i.e., Manual on Uniform Traffic Control Devices) is used primarily in the United States; and the SADC Convention (i.e., Southern African Development Community) is used primarily in Africa. Variations of these conventions may be defined and used for more specific geo-spatial regions. For example, each state in the United States may either adhere to the federal version of MUTCD or they may define their own state-specific version. Each road sign represented by RSO may be associated with the convention to which it adheres.

Conventions also provide standard images that depict the sign. These standard images are often referred to as prototypes and provide a template for the design, construction, and illustration of signs in manuals. Prototypes often come in two versions, a full color version and a gray-scale version. RSO enables road signs to link to these prototype images on the Web.    

\subsection{Alignment with Domain Vocabularies}
\label{sec-alignment}
While the Road Sign Ontology defines a general vocabulary to catalog visual and conventional properties 
of road signs, we need to extend it to include other non-ontological structure of existing road sign datasets, and to tailor to domain-specific (ADAS) applications used in commercial systems. This is important to allow the machine learning and computer vision models, which is developed using the specific labels of the datasets or domain-specific taxonomy, to interoperate with the knowledge graphs. 
In our work, we integrate three open datasets: LISA \cite{lisa2012}, representing US signs, GTSRB \cite{gtsrb2011}, representing German signs, and TT-100K \cite{zhu2016traffic}, representing Chinese signs. Each of these datasets have a vocabulary that is tailored to representing signs only in a specific region. In addition, we also model the taxonomy used in our different in-house annotation applications, which cover road signs of 72 countries, and are optimized for TSR function development. Below we present the high-level approach of our alignment.

To integrate these datasets, we create domain-specific vocabularies and put them under each corresponding namespace. 
Of course, there is a significant overlap between namespaces, but the alignment is non-trivial, for example in two following cases:

\textit{Road Sign Categories} — The way that road signs are categorized is different for each domain and dataset. Each uses a distinct number of categories with divergent levels of granularity. As an example, 
a road sign (e.g., \textit{truck speed limit 60}) can be mapped to a single category \textit{truck speed limit} within one dataset
(TT-100K), 
while mapped to both a \textit{truck} sign category and a \textit{speed limit} sign category in another (GTSRB).

\textit{Road Sign Features} — Road sign features are also represented at different levels of abstraction within the various domains and datasets. 
While our special in-house annotation tools enable annotators to describe detailed features of the road signs, open datasets only use prototypical images. Furthermore, GTSRB provides for each sign different prototypes with variants in foreground, background, and border, while TT-100k and LISA use only one prototype per each.

\begin{figure}[t!]
  \includegraphics[width=0.8\textwidth]{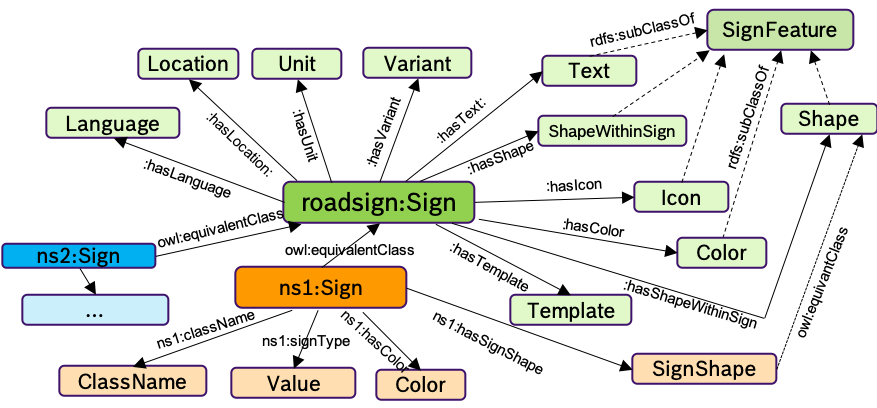}
  \caption{Example of aligning the concept of road sign, in RSO, with concepts in other vocabularies.}

  \label{fig:kg-mapping}
\end{figure}

Figure \ref{fig:kg-mapping} illustrates how the Road Sign Ontology is aligned with  domain vocabularies. The standard \texttt{owl:equivalentClass} is used to link the sign categories, and new relations are introduced for equivalent road sign features. We also introduce the class \texttt{Variant} to embrace the different prototypes as in the case of GTSRB.

\section{Road Sign Knowledge Graph}
\label{sec:rs-kg}
This section describes our approach to build the Road Sign Knowledge Graph at scale based on RSO discussed in Section \ref{sec-rso}.
We need to design the knowledge graph specifically to enable the recognition of road signs in different applications. 
However, building the knowledge graph manually is both time consuming and difficult due to lack of comprehensive domain knowledge of human annotators. In our work, we develop a two-step system: First, we rely on the crowd to construct the large-scale graphs with basic properties. Second, we align and extend the graphs to "fine-tune" to domain-specific data and vocabularies.
Below we detail our approach.

\subsection{Road Sign Knowledge Graph Construction with Crowdsourcing}
\label{sec:rsr-kg-crowdsource}
As the number of localized road signs is considerably large and requires many people who know different languages used in different locations,  
we leverage crowds recruited from three online crowd-sourcing marketplaces with global registered work-forces:
Amazon Mechanical Turk~\cite{mturk_url}
, Click Worker~\cite{clickworker_url}
and UpWork~\cite{upwork_url}.
We design two tasks in each platform:

\paragraph{Identifying Road Sign Templates for Various Countries}
In the first task, we ask the crowd to find a Web site that provides official road sign documents having templates of the road sign for a target country or state. We ask at least three crowd workers to get consensus on recommended sources. After identifying a source, manual template extraction is required if the identified resource does not support a separate image file format for each road sign template.  

\begin{figure}[t!]
  \includegraphics[width=0.9\textwidth]{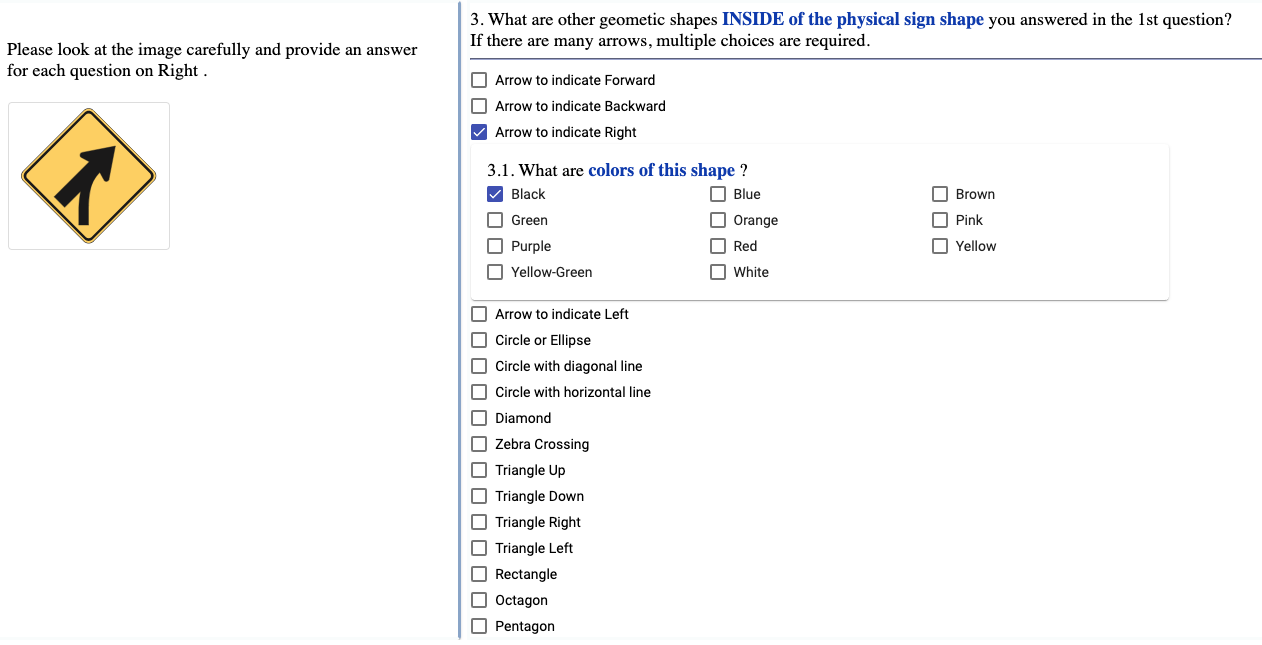}
  \caption{Screenshot of the Crowdsourcing Task for Road Sign Attribute Extraction}
  \label{fig:kg-annotation}
\end{figure}

\paragraph{Road Sign Feature Extraction} 
In the second task, we create a web-based application for crowd workers to extract the sign features used for constructing the road sign knowledge graph. A snapshot of the user interface is shown in Figure \ref{fig:kg-annotation}. In detail, the crowd worker is asked to look at a road sign template and provide the plate shape, background color, border color, additional shapes (e.g., left arrow) inside the plate, icons (e.g., vehicle), text and variants (e.g., street name) if applicable. This microtask can be done by any crowd workers and does not require road sign knowledge i.e., the meaning of a given road sign template. All answers can be selected from the provided options, except text that should be typed in the text field. Therefore, we do not specify any driving experience as qualification or require any training; we simply provide an instruction with examples and recruit workers whose approval rate is greather than 98\%.  However, each HIT includes one gold standard road sign, for which the system knows the ground truth in order to screen scammers who intentionally try to fool our system. Each individual sign template is presented to one worker, and one internal expert reviews the answers from the worker, followed by an additional review with another internal expert for further clarification if necessary.

We have constructed a generic road sign knowledge graph with 2,804 road signs from two different countries as of the reporting date. (The number of signs will be increased when adding signs for more countries.) In detail, 281 HITs (i.e., Human Intelligent Task) are published to Amazon Mechanical Turk when extracting sign features for U.S federal signs, three US states signs and German signs. Each HIT includes up to 10 road sign templates. The number of questions for each sign template varied based on sign features shown in each template. 7 to 11 questions are answered for an individual sign template. It took 23.7 minutes on average to complete one HIT. The workers provided correct answers (98\%) for most questions on shapes, icons, and text. We observed that some colors such as yellow and orange are not properly annotated by a handful workers, possibly due to color representation in the screen they used. We did not reject these answers but correct the answers during the review process. Surprisingly 35\% of answers are incorrect when they are asked to provide variants or units for certain signs. For instance, a no parking sign (see the last sign in \ref{fig:signs_variation}) includes a time variant (8:30 AM to 5:30PM), but some workers were not able to recognize this time variant. Also the ground truth for the SPEED LIMIT 30 unit is "miles per hour", but some workers gave "miles" as the unit. 41\% of workers reported that they drive everyday and 6\% of workers do not have a driving license. We have not seen any differences in the road sign knowledge graph annotation quality with respect to workers' driving experience. 1\% of scammers are detected when running this batch of HITs.   

Validated attributes for each sign template are translated into RDF facts of a corresponding entity of type \texttt{Sign} in the \emph{generic knowledge graph}. 
Next, we describe how the facts are refined to produce different domain-specific graphs. 

\subsection{Alignment of Road Sign Knowledge Graph to Domain-specific Knowledge Graphs}
To get the knowledge graphs specific to different domains, we first 
create a separate graph for each domain from the relevant sub-graph of the generic knowledge graph (for instance, a sub-graph containing all facts about a target country or state).
We then perform the following three alignments to extend and refine the domain-specific graphs:


\begin{itemize}
  \item \textbf{Automated Reasoning}: As our RSO is compliant to OWL-DL, the system can perform semantic reasoning to add more facts, such as adding category to the sign via its color and shapes. The reasoning can also create facts in different granularity .
  For example, the property \textit{has foreground color} can be refined to \textit{has icon color}, \textit{has text color}, \textit{has shape within a sign color} using the subsumption, or mapped to a more generic property \textit{has color}.
  
  \item \textbf{Auto-transformation for individual triples}: If the content in a triple in the generic graph is transformable, we apply rules to get more facts. For instance, the text "SPEED LIMIT 30" 
  can be transformed into two triples with "SPEED LIMIT" as a text and 30 as a numerical value in the domain-specific graph. 
  
  \item \textbf{Manual alignment}: Finally, experts of ontology alignment are also advised to add new vocabularies into the domain-specific graphs when needed. For example, the category/class names used in a domain are often acronyms, which cannot be automated without additional inputs. 
  
\end{itemize}

The result of this step is a generic road sign knowledge graph and multiple domain-specific graphs. The graphs with new signs are regularly reviewed by domain experts. Meanwhile, only relevant graphs are loaded into the downstream annotation task (described in Section~\ref{sec:system}) to mitigate implications of any remaining inconsistency.

\subsection{Notes on Architecture}
The knowledge graphs are stored and processed in the StarDog Enterprise Cluster\footnote{\url{https://www.stardog.com}}. We use a MongoDB\footnote{\url{https://www.mongodb.com}} database to store intermediate annotations and perform multiple validation before storing it in StarDog. We extended JenaBean \cite{cowan2008jenabean} to convert the Web application data model to the triples that follow Road Sign Ontology (RSO) described in Section \ref{sec-rso}. We use the built-in semantic reasoning and regular expression capabilities in StarDog to perform the graph alignment. 


\section{Knowledge Graph and Machine Learning Assisted Road Sign Annotation Pipeline}
\label{sec:system}
This section describes how the proposed Road Sign Knowledge Graph and the domain Road Sign Knowledge Graph is used in our road sign annotation tool, along with a machine learning classifier, to assist a human annotator on an image.  

 \begin{figure*}[t!]
  \includegraphics[width=0.9\textwidth]{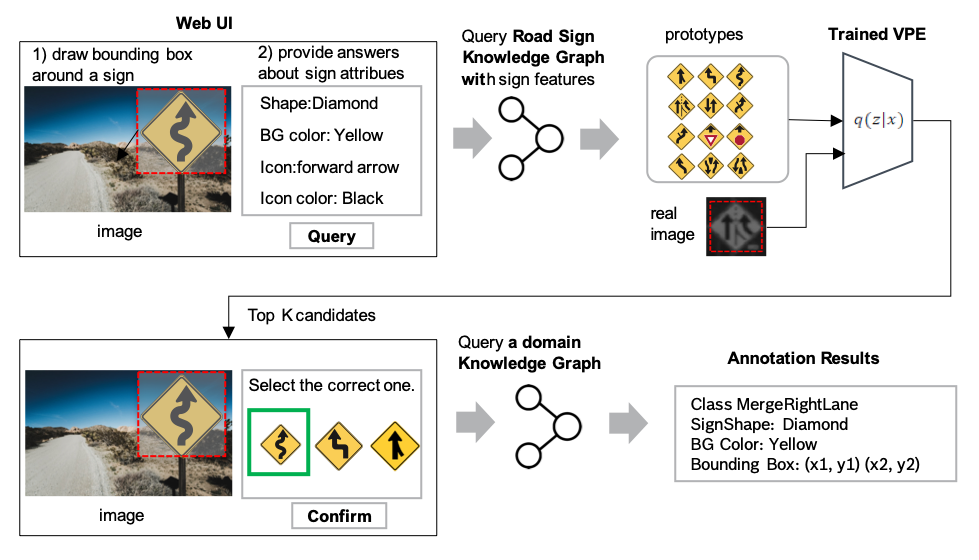}
  \caption{Road Sign Annotation Process}
  \label{fig:kg-process}
\end{figure*}


\paragraph{Tasks for the Human Annotator}
  As shown in Figure \ref{fig:kg-process}, the task for the human annotators is to draw a bounding box around a road sign on an image and select a matching sign prototype from a small palette of signs. We do not expect the human annotators possess knowledge on traffic systems which often vary across countries. Furthermore, this simple task does not require a separate training session about road signs to complete. Instead, annotators interact with the system by providing road sign features which are visible inside of the bounding box they draw on the Web UI. The annotator provides visual attributes such as plate sign shape and background color as common attributes, and icons, text and additional shapes as optional attributes. Then, they are asked to select a sign template from the ordered candidates. The task execution time and quality depends on the search space, i.e., the number of candidates they have to visually compare, and the quality of the image. If a road sign is not exactly matching with any candidates, then a worker selects a most matching sign or a default sign template that represents common attributes such as plate shape and background color with an indication of missing a sign template. This indication is further used by the system to add missing signs to the knowledge graph. Note that road signs in the wild are sometimes customized and knowledge graph constructed based on the official documents cannot cover all possible variation at the beginning of the system deployment. Rather we support to scale up while running the system with workers' feedback.     

\paragraph{Use of Road Sign Knowledge Graph and Machine Learning Model}
 Our tool supports the human annotator by providing a handful of road sign candidates that match the attributes given through a knowledge graph search. If the number of sign candidates is greater than a threshold \textit{K}, then the machine learning model is applied to further reduce the number of candidates. 
 
 We integrate the Variational Prototypical Encode (VPE) model \cite{vpe2019} to predict \textit{top-K} road sign candidates. The inputs for the VPE model are: 1) a cropped image patch around the bounding box that the annotator draws on the real road image and 2) sign templates filtered by the Road Sign Knowledge Graph. These two inputs are encoded into the latent space, and nearest neighbor classification is used to rank road sign templates. The system returns top \textit{K} candidates back to the human annotator.   
 
 Prediction of unseen classes is crucial in the road sign annotation due to rare road signs in the long tail. Model prediction of conventional classification with known classes cannot assist human annotators effectively unless that model is trained with large datasets that include rare classes. The encoder in the training phase of VPE encodes real images to latent distribution and the decoder reconstructs the encoded distribution back to a prototype that corresponds to the input image. By doing so, the trained encoder is used as a feature extractor and VPE learns image similarity and prototypical concepts instead of learning classification itself. Therefore, the pre-trained model can predict novel classes that are unknown during the training time.

\paragraph{Querying the Domain Knowledge Graph and Human Validation}
Upon the final selection of a matching road sign prototype, the system queries our domain specific knowledge graph to get relevant attributes needed for annotation, such as a corresponding class name, class description, colors, sign shape and text along with a coordinate of the bounding box.
\section{Evaluation} 

This section provides experimental results to validate our assumption that the knowledge graph and VPE model can reduce the number of sign templates which the annotator has to search through to create ground truth.


\subsection{Search Space Reduction with Road Sign Knowledge Graph}
First we discuss the search space reduction from the knowledge graph alone.

\paragraph{Dataset Used in the Evaluation} Our first evaluation uses 51,000 frames of road scenes recorded from one of front cameras mounted in our testing vehicles. The recorded frames cover various cities in US. A total of 6,253 road signs instances are observed. 50 road sign classes are annotated in this dataset. Note that a road sign class in this data model can have multiple sign templates if the meaning of signs are the same in the application. For example, SPEED LIMIT 30 and SPEED LIMIT 50 have the same road sign class with a different value (30 and 50 respectively) for each sign. Some road signs that are not relevant to the application are modeled as other classes.   \\    

\paragraph{Knowledge Graph Used in the Evaluation} The Road Sign Knowledge Graph constructed with 845 US federal MUTCD sign templates are used for this experiment. We select only one sign template for each class and run SPARQL queries. Common query parameters for all signs are plate shape and background color. Shape within a sign, icon and text are used optionally depending on the sign template. \\

\paragraph{Evaluation Results}
To understand the distribution of signs on the graph, we run queries with common parameters (shape and background color) on the graph database. The query results show that 42\% of signs have rectangle shape and white background color, and 14\% of signs have diamond shape and yellow background color. 

We run 50 queries with common and optional query parameters to evaluate the search space reduction for observed sign classes in our dataset. Table \ref{tab:kg_search_space} shows the distribution of the search space size range used from the evaluation. The results show that the size of the search space is reduced by 98.9\% (from 845 to a mean of 8.92 signs) on average when querying road signs with minimum 3 and maximum 5 attributes. The reduced number of road sign candidates should decrease the efforts for the human annotator because the number of comparison against a real sign is reduced proportionally. 

\begin{center}
\captionof{table}{Search space reduction distribution \label{tab:kg_search_space}}
\begin{tabular}{ |c|c|c|c| } 
\hline
 search space size & percentage(\%) & mean & stdev \\ 
 \hline
 1 to 5  & 38\% & \multirow{5}{*}{8.92} & \multirow{5}{*}{7.36}\\  
 \cline{1-2}
 6 to 10  & 16\% & &\\
 \cline{1-2}
 11 to 15 & 20\% & &\\
 \cline{1-2}
16 to 20 & 12\% & &\\
\cline{1-2} 
21 to 25 & 14\% & &\\
\hline
\end{tabular}
\end{center}

\paragraph{Training Time and Learning Curve} 
In addition, human annotators spend weeks to get trained on road sign annotation for various countries to be able to make intelligent decisions when they annotate signs that are different from provided road sign templates, according to our interviews with annotation experts. Uncertainty in classifying variants of similar road signs creates either failure in the ground truth annotation or requires additional rigorous reviews relying on the trained domain expert. Training time and learning curve can be improved by increasing the number of road sign prototypes via crowd-sourced knowledge graph construction. 

\subsection{Variational Prototyping-Encoder Model Accuracy with Respect to Search Space Size}
The VPE re-ranks the sign candidates filtered by the knowledge graph. Our second hypothesis is that the accuracy of VPE can be improved with the reduced number of candidate classes to the encoder during inference. This improvement can further increase the efficiency of the human annotators by further reducing the search space. \\


\paragraph{Datasets used in the Evaluation} We validate this hypothesis on two benchmark road sign datasets to evaluate how the reduction of the search space from knowledge graph improves the classification accuracy: 1) German Traffic Sign Recognition Benchmark (GTSRB)\cite{gtsrb2011}, containing 43 classes of road signs used in Germany where they adopt the Vienna convention. This dataset provides approximately 50,000 instances of image patches with sizes varying from 15x15 pixels to 250x250 pixels. The dataset also provides road sign template images. 2) LISA\cite{lisa2012} dataset, containing 47 US sign classes of 7,855 annotation from the Manual on Uniform Traffic Control Devices (MUTCD) convention. The LISA dataset does not contain the road sign template images; we used the road sign templates from our Road Sign Knowledge Graph. \\

\paragraph{Training and Testing Process}
We selected these datasets because significant differences in signs between these two countries allow us to evaluate the effectiveness of our knowledge graph in the VPE model when testing the model performance with road sign classes not used during the training phase. We first trained VPE with the GTSRB dataset. The input of the VPE model is cropped image patches from the real road scenes and the output of the VPE model is the road sign template for the correct class. The VPE learned image similarity represented in embedding (vector size of 300) as well as prototypical concepts of the road sign during the training phase. 

10 signs classes having the diamond shape and the yellow background color in the LISA dataset are used to test the performance by changing the number of road sign templates to the pre-trained encoder. No signs in GTSRB have the diamond shape and yellow background color. The performance covers unseen sign classes with different features during the training time. The output of the pre-trained encoder is the ranked list of road sign templates. We calculated accuracy of top 1, top 3, and top 5 ranks (the ground truth template is listed in top 1, top 3, and top 5 ranks). \\

\paragraph{Evaluation Results} The evaluation results shows that the accuracy of the model is increased by 10-13\% when the search space size is reduced from 30 to 10 (See Table \ref{tab:vpe}).  The results also demonstrate that the model predicts the ground truth template for 75\% of the dataset when the search space size is 10. In these cases, the human annotator would not need to spend any time searching for a matching sign, thus further reducing annotation effort.


\begin{center}
\captionof{table}{VPE Performance Evaluation with Different 
Search Spaces Tested with Signs having Diamond Shape and Yellow Background Color\label{tab:vpe}}
\begin{adjustbox}{width=0.9\textwidth,center}
\begin{tabular}{ |c|c|c|c|c| }
\hline
 search space size s & accuracy (top-1) & accuracy \newline (top-3) &accuracy \newline (top-5) \\ 
 \hline
 10 & 0.73 & 0.85 & 0.90  \\  
 \hline
 20 & 0.69 & 0.80 & 0.85 \\
 \hline
 30 & 0.60 & 0.73 & 0.80 \\
 \hline
\end{tabular}
\end{adjustbox}
\end{center}
\section{Conclusion} 

We proposed to use Road Sign Knowledge Graph for the data annotation tool for road sign annotation along with a meta-learning model. The ontology model for the road signs is intended to map to various domains and applications. We demonstrate that the construction of the Road Sign Knowledge Graph can be made scalable for signs across states and countries by using a crowd-sourcing approach. The evaluation results show that the search space for human annotators can be reduced by 98.9\%, from a palette of 845 signs to only 8.92 signs, reducing human effort and easing the learning curve for road sign classification. Furthermore, the results show that integrating machine learning with the knowledge graph can produce an exact match for 75\% of signs, eliminating human classification effort entirely in those cases.  

\bibliographystyle{plain}
\bibliography{refs}
\end{document}